\documentclass[conference]{IEEEtran}
\usepackage{cite, epsfig, mathptmx, times, amssymb, amsmath, color, subfigure}
\usepackage{subfigure,graphicx,booktabs,soul,color,graphics,epsfig,mathptmx,moreverb,ifpdf,cancel,xcolor,amsthm,framed,dblfloatfix}
\usepackage{gensymb,graphicx,url, multirow,kantlipsum}
\usepackage{fancyhdr}
\usepackage{placeins}
\usepackage{balance}
\usepackage{flushend}

\usepackage{siunitx}

\usepackage{algorithm}
\usepackage[noend]{algpseudocode}
\usepackage{hhline}
\usepackage{cellspace} 

\usepackage{makecell} 
\setcellgapes{1.75pt}

\algrenewcommand\algorithmicforall{\textbf{foreach}}
\algrenewcommand\algorithmicindent{.8em}

\ifCLASSINFOpdf
\else
\fi
\makeatletter
\def\ps@IEEEtitlepagestyle{%
	\def\@oddfoot{\mycopyrightnotice}%
	\def\@evenfoot{}%
}
\def\mycopyrightnotice{%
	{ \footnotesize 978-1-5090-6173-0/16/\$31.00 ~ \copyright 2017 IEEE\hfill}
}
\makeatother

\DeclareRobustCommand*{\IEEEauthorrefmark}[1]{%
	\raisebox{0pt}[0pt][0pt]{\textsuperscript{\footnotesize\ensuremath{#1}}}}

\usepackage{xcolor,colortbl}
\definecolor{ColorParamTaken}{rgb}{0.7,1,0.7}
\definecolor{ColorParamConst}{rgb}{0.88,1,1}
\definecolor{ColorParamNeglected}{rgb}{1,0.7,0.7}

\title{\LARGE \bf
UAV Control in Close Proximities - Ceiling Effect on Battery Lifetime}

\author{\IEEEauthorblockN{\IEEEauthorblockN{
			Basaran Bahadir Kocer\IEEEauthorrefmark{1,4},
			Volkan Kumtepeli\IEEEauthorrefmark{2},
			Tegoeh Tjahjowidodo\IEEEauthorrefmark{1},
			Mahardhika Pratama\IEEEauthorrefmark{3}, \\
			Anshuman Tripathi\IEEEauthorrefmark{4},
			Gerald Seet Gim Lee\IEEEauthorrefmark{1},
			Youyi Wang\IEEEauthorrefmark{5}}
		\IEEEauthorblockA{
		    \IEEEauthorrefmark{1}School of Mechanical and Aerospace Engineering, Nanyang Technological University, Singapore \\            
			\IEEEauthorrefmark{2}Energy Research Institute @ NTU, Interdisciplinary Graduate School, Nanyang Technological University, Singapore \\
			\IEEEauthorrefmark{3}School of Computer Science and Engineering, Nanyang Technological University, Singapore   \\
			\IEEEauthorrefmark{4}Energy Research Institute @ NTU, Nanyang Technological University, Singapore   \\
			\IEEEauthorrefmark{5}School of Electrical \& Electronic Engineering, Nanyang Technological University, Singapore }}}

\begin{document}

\maketitle
\thispagestyle{empty}
\pagestyle{empty}

%%%%%%%%%%%%%%%%%%%%%%%%%%%%%%%%%%%%%%%%%%%%%%%%%%%%%%%%%%%%%%%%%%%%%%%%%%%%%%%%
\begin{abstract}
	With the recent developments in the unmanned aerial vehicles (UAV), it is expected them to interact and collaborate with their surrounding objects, other robots and people in order to wisely plan and execute particular tasks. Although these interaction operations are inherently challenging as compared to free-flight missions, they might bring diverse advantages. One of them is their basic aerodynamic interaction during the flight in close proximities which can result in a reduction of the controller effort. In this study, by collecting real-time data, we have observed that the current drawn by the battery can be decreased while flying very close to the surroundings with the help of the ceiling effect. For the first time, this phenomenon is analyzed in terms of battery lifetime degradation by using a simple full equivalent cycle counting method. Results show that cycling related effect on battery degradation can be reduced by a 15.77\% if the UAV can utilize ceiling effect.  
	
\end{abstract}
\begin{IEEEkeywords}
UAV, model predictive control, battery lifetime, energy efficiency, ceiling effect, ground effect, interaction control, longer flight time.
\end{IEEEkeywords}

%%%%%%%%%%%%%%%%%%%%%%%%%%%%%%%%%%%%%%%%%%%%%%%%%%%%%%%%%%%%%%%%%%%%%%%%%%%%%%%%
\section{Introduction}

With the developments in the sensors \cite{koccer2013development}, computational tools \cite{houska2011acado}, and hardware technology \cite{ruggiero2018aerial}, the control of the UAV in free-flight became a mature subject. Nowadays, the expectation of the UAV flight pushes the operation limits towards interaction with its environment, where the UAV collaborates with people, different robots and surrounding objects \cite{kocer_2018a,kocer_2018b,kocer_icarcv}. It is explored that the thrust characteristics of the UAV change when it interacts with an object, particularly in close proximities \cite{ceiling2018_iros,conyers2018aempirical,conyers2018bempirical}.

The ceiling effect, known as altered rotor characteristics while the UAV flies very close to the ceiling, is investigated in \cite{kocer_2018_ceiling}  where the system is identified for certain points below the ceiling. It has been seen that since the rotor speed and the thrust efficiency increase at the same time while the system operates within 10 centimeters below the ceiling, the UAV consumes less energy to stay and the locomote by leveraging the ceiling effect.

The capacity of the batteries is one of the main bottlenecks of the UAV operations, which poses an obstacle to achieving longer flight times. In order to extend the flight time, it is possible to decrease weights on the structure and/or replace the motors with the higher power-to-weight ratio ones. However, these solutions are trivial and do not promote any technological breakthroughs. Therefore, there is a need for operational planning considering algorithmic and/or aerodynamic aspects which can also pave the way for exploration of control methods that leverage advanced hybrid storage system which has high energy and high power densities at the same time. Although hybrid storage technology is more mature in the ground vehicles \cite{vadlamudi2016hybrid}, there are also some initial studies for the aerial systems such as the work in \cite{edwards2018integrating} on the usage of hydrogen fuel cells on UAVs. 

In order to extend overall flight time for the UAVs, an optimal route can be generated \cite{morbidi2016minimum} by considering minimum energy consumption by the motors. This study is further extended in \cite{morbidi2018energy} for a tilting propeller system. A similar approach is proposed in \cite{vicencio2015energy,gandolfo2017stable} without considering the physical motor parameters. Except for the efficient path planning approaches, one of the techniques is to consider a physical modification to exclude the exhausted battery modules to reduce the total weight \cite{chang2015improving}. In order to extend the total mission time, it is also possible to use the power over tether concept for the UAV \cite{nicotra2017nonlinear}. Similarly, use of wireless power transfer system to minimize downtime for recharging is also implemented in \cite{junaid2016design,plaizier2018design}. Moreover, some studies consider the estimation of the endurance when planning the path \cite{abdilla2015power,gatti2015maximum}. Although the literature appears to contain some work for increasing flight range by using less energy from the battery, none of the studies try to increase efficient use of battery nor analyze the effect of their algorithm on the battery degradation. 

As being different from the previous studies, a battery lifetime analysis is proposed by exploiting the efficient flight in close proximities, i.e., under the ceiling effect. For the first time, we present the battery analysis for the lifelong operation of the UAV. Since there is a need to frequently charge and discharge the UAV battery, the cycling effect, as well as the depth of discharge, are considered. By collecting the online battery state during the flight, the battery longevity has been analyzed by taking the cycling effect into account. Additionally, the increase on the flight time is also discussed. 

\begin{figure*}[b!]
	\centering
	\includegraphics[height=2in]{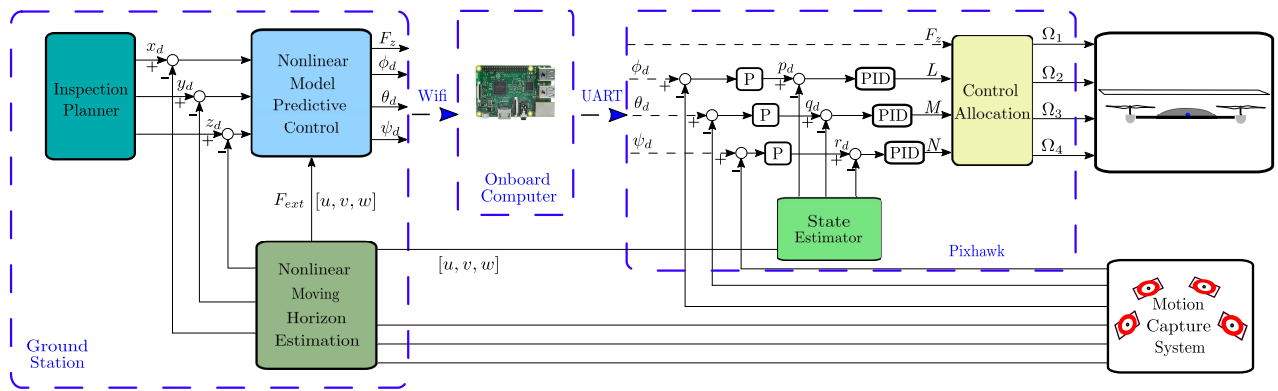}
	\caption{UAV interaction scheme: estimation and control structure}
	\label{fig:control_block}
\end{figure*}

This paper is organized as follows: The ceiling effect in close proximities is presented in Section II. The considered battery lifetime model for the UAVs is introduced in Section III. The test scenario of the aging for the UAV batteries and numerical investigations are introduced and discussed in Section IV. Finally, conclusions including future works are drawn in Section V.

\section{Ceiling Effect}

\begin{figure}[b!]
	\centering
	\includegraphics[width=0.4\textwidth]{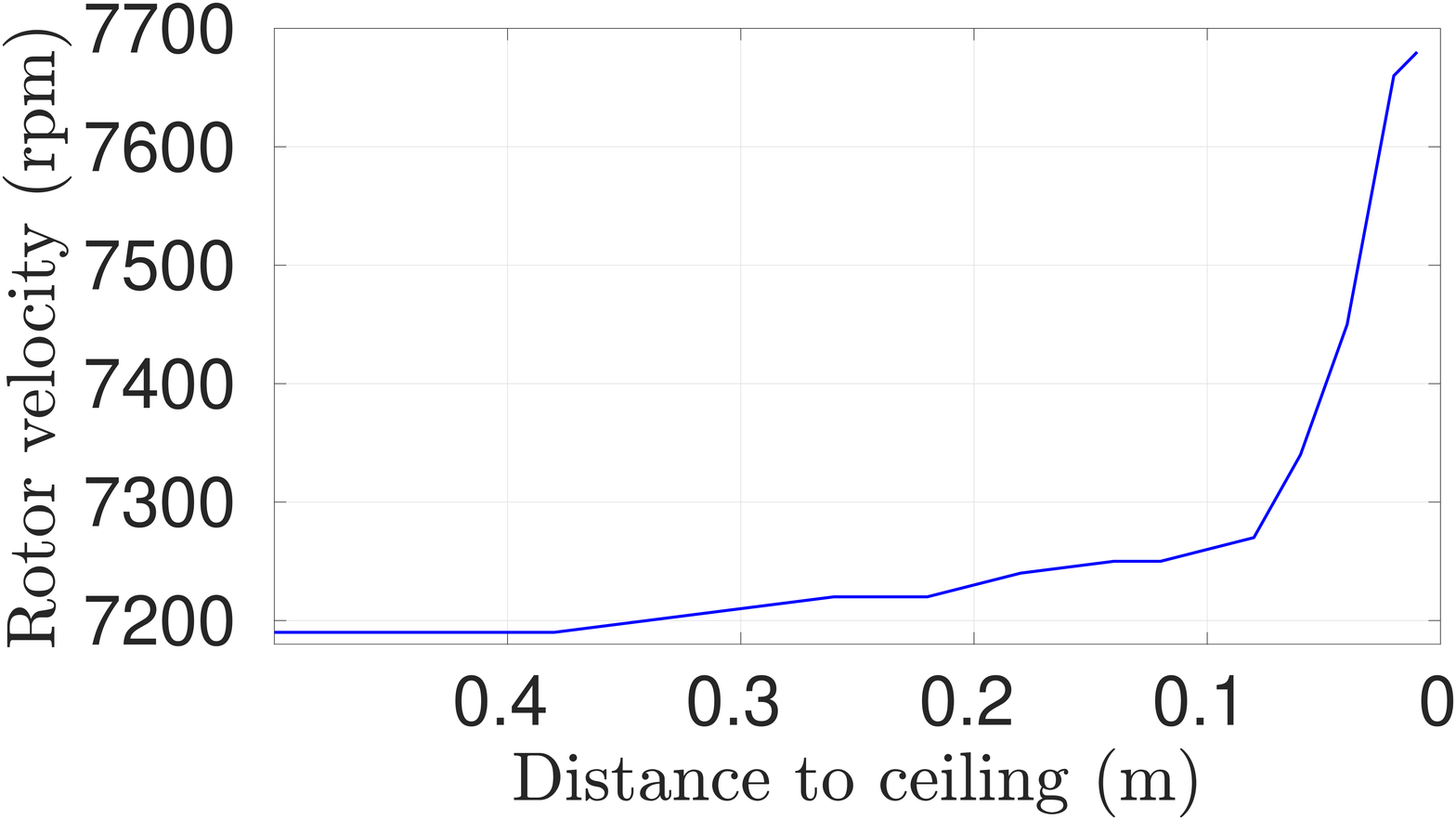}
	\caption{UAV rotor speed characteristics in close proximities \cite{kocer_2018_ceiling}}
	\label{fig:ceilingrpm}
\end{figure}

\begin{figure}[b!]
	\centering
	\includegraphics[width=0.4\textwidth]{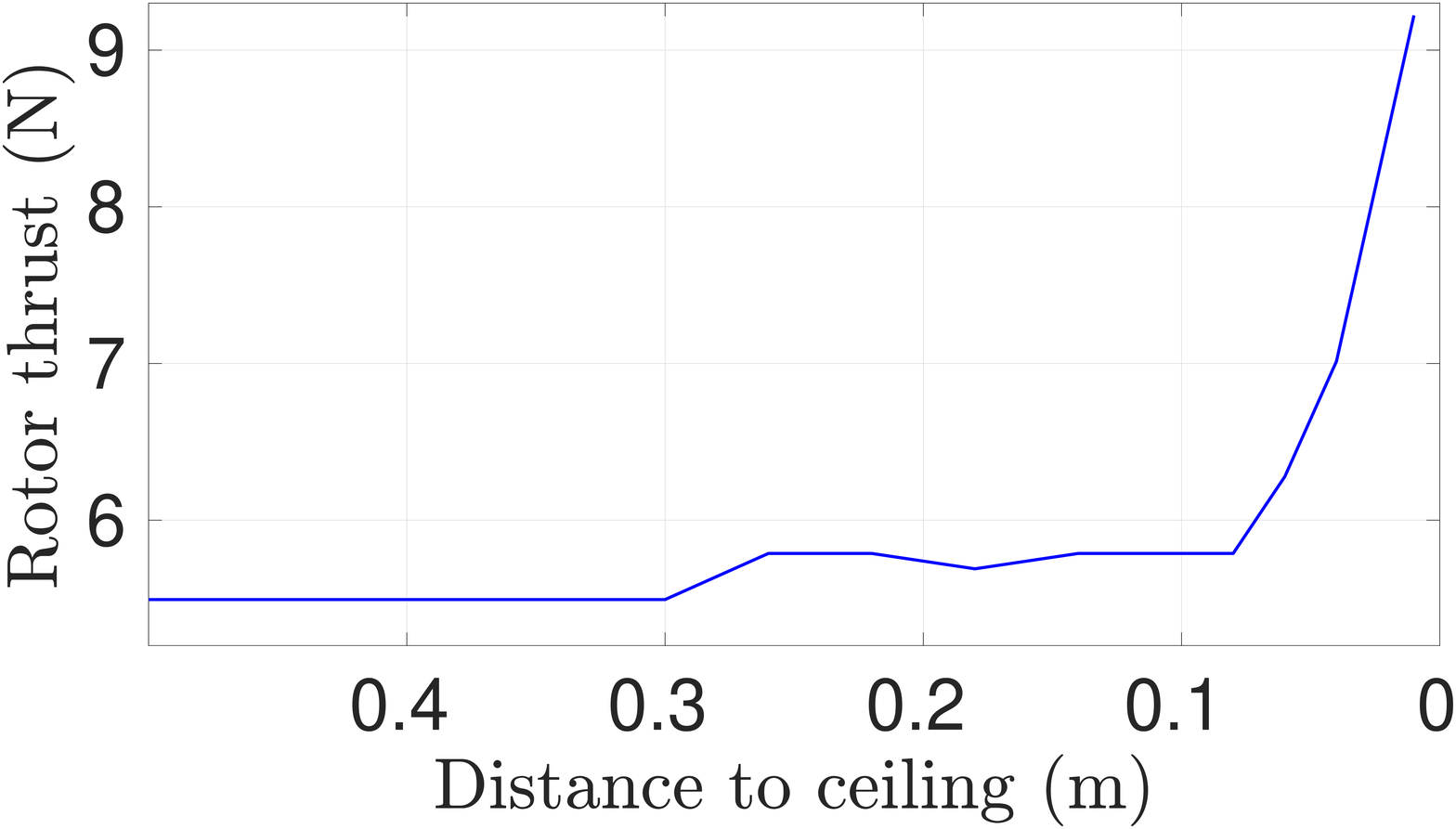}
	\caption{UAV thrust characteristics in close proximities \cite{kocer_2018_ceiling}}
	\label{fig:ceilingthrust}
\end{figure}

While the UAV flies in close proximities under the surrounding environment, the flow at the top the UAV increases the rotor wake. Similar to the vacuum effect, this phenomenon pushes the UAV upward. In addition, there is a significant increase in the rotor speed and the thrust efficiency at the same time. This effect is illustrated in Fig. \ref{fig:ceilingrpm} and \ref{fig:ceilingthrust}. By keeping the throttle command constant, it is observed that the rotor speed increases while the gap between the rotor and ceiling decreases. At the same time, the thrust efficiency increases while the rotor approaches the ceiling. Without a proper control strategy, the system might crash during these instants. Fortunately, we have developed a control strategy consisting of force estimation and nonlinear model predictive controller. This scheme handles this complex interaction to conduct the predefined operation for the UAV. The block diagram may be seen in Fig. \ref{fig:control_block}.  This challenging phase reduces the energy spent for the UAV to stay and/or locomote under the ceiling effect. 


\section{Battery Lifetime Analysis}

After implementing the developed force feedback control algorithm, UAV was able to fly very close to the ceiling that is enough to see a considerable ceiling effect. Although our aim was to design a controller which can help UAV to fly under strong disturbances; we have observed that ceiling effect can be utilized by the UAV to reduce control effort. In these experiments, the current drawn from the battery is reduced from the range of 9.5 A to 8 A range in the vicinity of the ceiling. Therefore, by considering the positive effect of decreasing current on the battery; we have conducted a simple battery-life analysis. According to \cite{schmalstieg2014holistic}, degradation in battery life, $\Delta SOH$, can be written in the superposition of two aging factors as calendar and cycle aging. Since we do not have a reliable battery aging model for Li-Polymer (LiPo) battery used in our UAV, it is assumed that batteries used UAVs are exposed to frequent charging and discharging. This assumption makes the cycling effect dominant. Therefore; calendar aging effects which mostly depend on the state of charge (SOC), temperature and time \cite{ecker2012development,broussely2001aging} can be neglected. Although there are different effects on the cycle aging, we would like to relate our cycle aging with the number of times the battery has been cycled. Consequently, we assume that our battery degradation will be proportional to the number of full equivalent cycles (FEC) as given in Eq. \eqref{eq:app:dSOHprop}.
\begin{eqnarray}
\Delta SOH \propto FEC \label{eq:app:dSOHprop},
\end{eqnarray}
where a unit $FEC$ is defined by fully charging and discharging the battery. It is updated at battery mode changes such as transitions between charge, discharge or idle. To explain calculation methodology, let us say that our mode starts at time $t_1$ and ends at $t_2$. Then,  the mean C-rate in this interval is given by \eqref{eq:Cr} and \eqref{eq:Crt}.
\begin{eqnarray}
\label{eq:Cr}
C_{r} &=& \frac{1}{t_{2} - t_{1}}\int_{t_{1}}^{t_{2}}C_{r}(t)dt,\\
\label{eq:Crt}
C_{r}(t) &=&  \frac{|i(t)|}{C_{nom}},
\end{eqnarray}
where $C_{r}$ and $i(t)$ are average C-rate and current drawn from battery at time $t$. $C_{nom}$ denotes the nominal capacity of battery in [Ah]. Then, the change in $FEC$ can be given by \eqref{eq:dFEC} and $FEC$ and $t_2$ can be written as in \eqref{eq:FEC1},
\begin{eqnarray}
\label{eq:dFEC}
\Delta FEC &=& \frac{1}{2} \cdot C_{r} \cdot (t_2 - t_1), \\
\label{eq:FEC1}
FEC(t_2) &=& FEC(t_1) + \Delta FEC, 
\end{eqnarray}
or equivalently in \eqref{eq:FEC2},
\begin{eqnarray}
\label{eq:FEC2}
FEC(t) &=& \frac{1}{2\cdot C_{nom}} \int_{0}^{t}|i(t)| dt.
\end{eqnarray}

Although it does not make any difference in our case, discrete update equation given in \eqref{eq:FEC1} is used aging calculations due to non-linear relation between aging and influence factors. Therefore, we prefer using \eqref{eq:FEC1} to create a base for the potential extensions in the future. Lastly, SOC is given by \eqref{eq:dSOC},
\begin{eqnarray}
\label{eq:dSOC}
SOC(t) = \int_{0}^{t} \frac{i(t)}{C_{nom} \cdot SOH(t)}dt.
\end{eqnarray}
Since we do not have the exact aging information; it is assumed that the state of health, $SOH$, in \eqref{eq:dSOC} is constant at 1. Therefore, $SOC$ equation becomes \eqref{eq:dSOC2},
\begin{eqnarray}
\label{eq:dSOC2}
SOC(t) = \int_{0}^{t} \frac{i(t)}{C_{nom} }dt.
\end{eqnarray}

\section{Test for Battery Aging}

\begin{figure}[b!]
	\centering
	\includegraphics[width=0.5\textwidth]{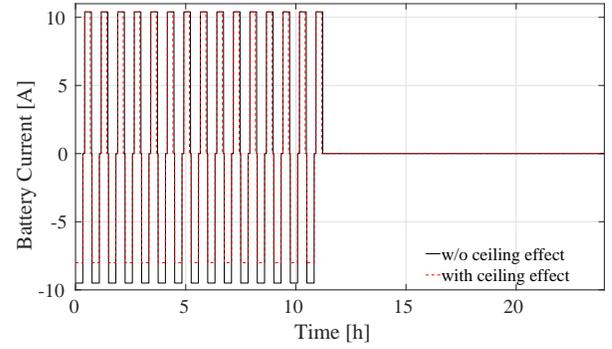}
	\caption{UAV load profile for one mission}
	\label{fig:dailyload}
\end{figure}

After introducing the methodology, we would like to create a test scenario to analyze the battery longevity. Let us say that we have an inspection mission in which the UAV does 15 flights with the endurance time of 20 minutes in each mission. During all operations, it starts from 95\% SOC, flies over 20 minutes, stays idle for 5 minutes then it is charged back to 95\% SOC in 20 minutes of time. This mission is repeated 15 times a day and then the UAV is unused until the next day. Self-discharge is neglected between days. When we apply this pattern to both cases with ceiling effect and without the effect, one-day load profile is given in Fig. \ref{fig:dailyload}. 

\begin{figure}[b!]
	\centering
	\includegraphics[width=0.5\textwidth]{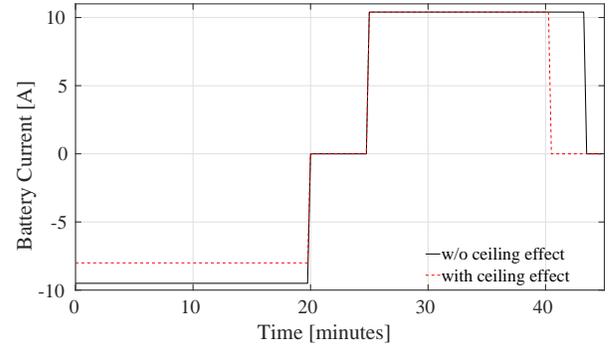}
	\caption{UAV daily load profile}
	\label{fig:eps_mission_day_current}
\end{figure}

\begin{figure}[b!]
	\centering
	\includegraphics[width=0.5\textwidth]{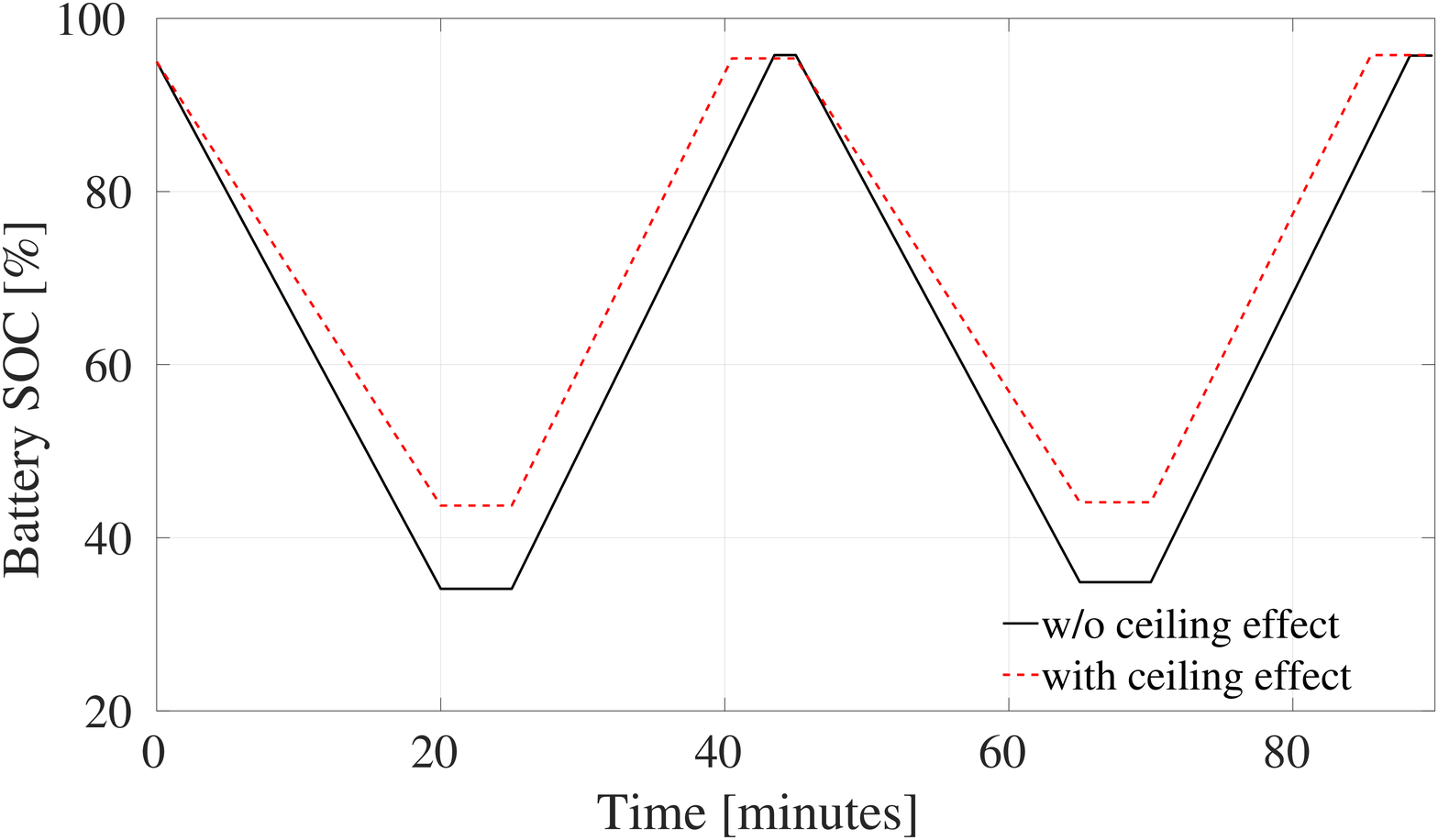}
	\caption{SOC change during two missions of UAV}
	\label{fig:eps_mission_day_SOC}
\end{figure}

It may be seen in Fig. \ref{fig:dailyload} that there is a load curve repeated 15 times for each mission then UAV is switched off. To see the curve clearly, we may zoom into one mission given in Fig. \ref{fig:eps_mission_day_current}.
Here it is seen that for the first 20 minutes, the UAV is flying and the battery is discharging. Then it rests for 5 minutes between 20th and 25th minutes. Then, it is charged for 20 minutes or until it reaches the necessary SOC. It should be noted that in the ceiling effect case, the battery is discharged with 8 A instead of 9.5 A as in the other case. Therefore, it loses less charge and it takes less time for it to charge back again. Although we use a charging current of 10.4 A (or 2C of C-rate), it finishes charging at 40.5 minutes while it takes until 43.5 minutes without ceiling effect, causing 3 minutes of difference. One major advantage of this situation would be increasing flight durations. Then we can see its effect on SOC in Fig. \ref{fig:eps_mission_day_SOC}.

In Fig. \ref{fig:eps_mission_day_SOC}, two missions are shown and it can be seen that SOC drops down to 43.7\% in one case and 34.1\% in the other case. Then it is increased to 95\% in both cases. When both cases are used for the same duration of the flight, ceiling effect causes less depth of discharge and can prolong battery life. 

\begin{figure}[b!]
	\centering
	\includegraphics[width=0.5\textwidth]{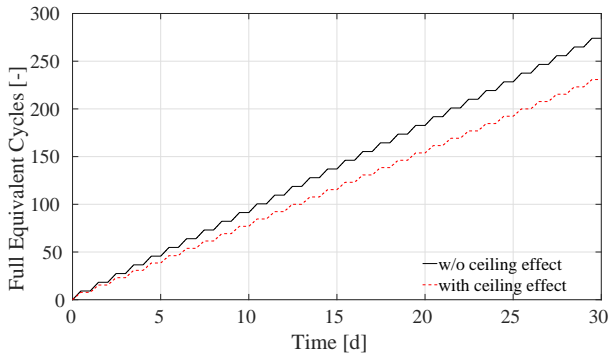}
	\caption{Battery full equivalent cycle in a month}
	\label{fig:eps_one_month_FEC}
\end{figure}

\begin{figure}[b!]
	\centering
	\includegraphics[width=0.5\textwidth]{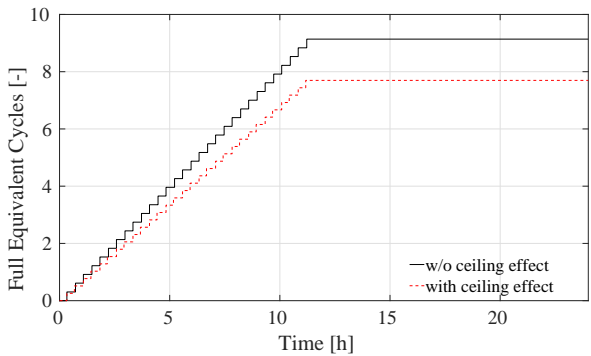}
	\caption{UAV load profile in hours}
	\label{fig:eps_one_day_FEC}
\end{figure}

Lastly, we would like to look at FEC since they directly affect battery life. When we look at the Fig. \ref{fig:eps_one_month_FEC}, we see that the number of cycles we do in different cases diverge. After a month, we cycle our battery 274 times without the ceiling effect whereas it is only 230 when we utilize the ceiling effect. With the help of force feedback, we reduce cycling effect 15.77\%. Since the battery degradation is directly affected by the cycling effect, we may say that degradation is also decreased in proportion to the FEC.   
In Fig. \ref{fig:eps_one_day_FEC}, we zoom into the FEC figure and see that FEC is updated discretely. Since we did not include the aging effect, the decrease from 9.14 to 7.7 is very close to the decrease of 30 days. 

\section{Conclusion and Future Work}

In this study, we have analyzed the effect of two different cases on the flight time and the battery longevity: (i) without ceiling effect; (ii) with ceiling effect. In the first case, the UAV needs to spend more energy which resulted in less endurance and the shorter lifetime of the battery. In the second case, the aerodynamic forces contribute the flight and the system spent less energy. In this case, the cycling effect is reduced by 15.77\%. At the same time, this causes the decrease in depth of discharge which also contributes the battery longevity although its effect is not calculated. Furthermore, the total operation time can be higher also when the system exploits the aerodynamic forces with the help of the ceiling effect. 

Since this study explores the reduction of the controller effort and its effect on the battery life of the UAV with assumptions, the following items can be considered in future studies:

\begin{itemize}
	\item  Instead of assuming a constant load current, a better flight profile that considers different modes from taking off to the landing can be utilized within the analysis.
	\item  A detailed aging model for the LiPo batteries can be used for the analysis to increase the accuracy. Having an aging model might also lead to an aging-aware controller design for the aerial missions, whereas aging-aware design concept is adopted by some power system related studies such as \cite{kumtepeli2019,goebel_2017}.
	\item Using a complete aging model also helps us to accurately calculate capacity degradation along with the state of health. Hence, capacity fade can be also accounted for a better estimation as in  \cite{hesse2017economic}. This would also allow us to build more resilient path planning algorithms. 
	\item For the experimental work, current values for different set of missions can be validated on the real batteries.  
	\item There is a need to release benchmark path profiles for a precise analysis for the battery.
	\item The same analysis can be conducted for the ground effect.
\end{itemize}

\section*{Acknowledgement}

The authors wish to thank for conducting the research work with support from the Energy Research Institute @ NTU (ERI@N) as well as the SINGA research scholarship.

%%%%%%%%%%%%%%%%%%%%%%%%%%%%%%%%%%%%%%%%%%%%%%%%%%%%%%%%%%%%%%%%%%%%%%%%%%%%%%%%
%%%%%%%%%%%%%%%%%%%%%%%%%%%%%%%%%%%%%%%%%%%%%%%%%%%%%%%%%%%%%%%%%%%%%%%%%%%%%%%%
%%%%%%%%%%%%%%%%%%%%%%%%%%%%%%%%%%%%%%%%%%%%%%%%%%%%%%%%%%%%%%%%%%%%%%%%%%%%%%%%
%\section*{APPENDIX}
%\section*{ACKNOWLEDGMENT}

%The preferred spelling of the word �acknowledgment� in America is without an �e� after the �g�. Avoid the stilted expression, %�One of us (R. B. G.) thanks . . .�  Instead, try �R. B. G. thanks�. Put sponsor acknowledgments in the unnumbered footnote on the %first page.

%%%%%%%%%%%%%%%%%%%%%%%%%%%%%%%%%%%%%%%%%%%%%%%%%%%%%%%%%%%%%%%%%%%%%%%%%%%%%%%%

%References are important to the reader; therefore, each citation must be complete and correct. If at all possible, references should be %commonly available publications.
\balance
\bibliography{References/bib}
\bibliographystyle{References/IEEEtran}
\end{document}